\documentclass{article}
\usepackage{geometry}
\usepackage[utf8]{inputenc}
\usepackage{amssymb}
\usepackage{amsmath}
\usepackage[ruled,vlined]{algorithm2e}
\usepackage{float}
\usepackage{longtable,booktabs}
\usepackage{graphicx}
\usepackage{hyperref}
\usepackage{cite}
\usepackage{multicol}
\usepackage{authblk}
\usepackage{enumitem}
\usepackage[noend]{algpseudocode}

\usepackage{adjustbox}
\usepackage{graphicx}
\usepackage[utf8]{inputenc}             
\usepackage[english]{babel}
\usepackage{tikz}
  \usetikzlibrary{shapes,arrows,fit,calc,positioning, decorations.pathreplacing}
  \tikzset{box/.style={draw, diamond, thick, text centered, minimum height=0.5cm, minimum width=1cm}}
  \tikzset{line/.style={draw, thick, -latex'}}
\usepackage{verbatim}
\usepackage{caption}
\graphicspath{ {./figure/} }

\title{Analysis and classification of main risk factors causing stroke in Shanxi Province \footnote{Correspondence:  zhouxiaoshuang@sxmu.edu.cn, hhuang@uic.edu.cn} \footnote{J. Liu and Y. Sun contributed equally to this work.} }

\author[a]{Junjie Liu}
\author[b]{Yiyang Sun}
\author[c,e]{Jing Ma}
\author[a]{Jiachen Tu}
\author[a]{Yuhui Deng}
\author[a]{Ping He}
\author[d,a,e]{Huaxiong Huang} 
\author[f]{Xiaoshuang Zhou}
\author[c]{Shixin Xu}
\affil[a]{BNU-
HKBU United International College, Zhuhai, China}
\affil[b]{Duke Kunshan University, 8 Duke Ave, Kunshan, Jiangsu, China}
\affil[c]{Laboratory of Mathematics and Complex Systems (Ministry of Education), 
School of Mathematical Sciences, 
Beijing Normal University, Beijing 100875, China}
\affil[d]{Department of Mathematics and Statistics, York University, Toronto, Ontario, Canada}

\affil[e]{Research Center for Mathematics,
Beijing Normal University at Zhuhai, 519087, China}
\affil[e]{Department of Mathematics and Statistics, York University,
Toronto, Ontario, Canada} 
\affil[f]{Department of Nephrology, Shanxi Provincial People’s Hospital, Taiyuan, Shanxi, China;}
\date{}

\begin{document}

\maketitle

\textbf{Abstract}

\textbf{Background}: In China, stroke  is the first leading cause of death in recent years.  It  is a major cause of long-term physical and cognitive impairment, which bring great pressure on National Public Health System.  Evaluation of the risk of getting stroke is important for the prevention and treatment of stroke in China.

\textbf{Methods}: A data set with 2000 hospitalized stroke patients in 2018 and 27583 residents during the year 2017 to 2020 is analyzed in this study. Due to data incompleteness, inconsistency and non-structured formats, missing values in the raw data  are filled with -1 as an abnormal class. With the cleaned features, three models on risk levels of getting stroke are built by using machine learning methods. The importance of ``8+2'' factors from China National Stroke Prevention Project (CSPP) is evaluated via decision tree and random forest models. Except ``8+2'' factors the importance of features and SHAP\footnote{SHAP: SHapley Additive exPlanations} values for lifestyle information, demographic information and medical measurement are evaluated and ranked via random forest model. Furthermore, a logistic regression model is applied to evaluate the probability of getting stroke for different risk levels.

\textbf{Results}: The risk of getting stroke for the 27583 residents is categorized and labeled with the CSPP's taxonomy: Low risk(11739 residents), Medium risk(7630 residents) and High risk(8214 residents). Among all ``8+2'' factors for the risk level of getting stroke, the decision tree model shows that top three features are  Hypertension (importance: $0.4995$), Physical Inactivity (importance: $0.08486$) and Diabetes Mellitus (importance: $0.07889$), and the random forest model shows that top three features are Hypertension (importance: $0.3966$), Hyperlipidemia (importance: $0.1229$) and Physical Inactivity (importance: $0.1146$). Except ``8+2'' factors the importance of features for lifestyle information, demographic information and medical measurement is evaluated via random forest model. It shows that top five features are Systolic blood pressure (importance: $0.3670$), Diastolic blood pressure (importance: $0.1541$), physical inactivity (importance: $0.0904$), Body Mass Index(BMI) (importance: $0.0721$) and Fasting Blood Glucose(FBG) (importance: $0.0531$). SHAP values show that Diastolic blood pressure, physical inactivity, Systolic blood pressure, BMI, smoking, FBG, and Triglyceride(TG) are positively correlated to the risk of getting stroke. High-density lipoprotein (HDL) is negatively correlated to the risk of getting stroke. Combining with the data of 2000 hospitalized stroke patients, the logistic regression model shows that the average probabilities of getting stroke are $7.20\%$ (with $95\%$ CI\footnote{CI: Confidence Interval}: $(6.65\%, 7.74\%)$) for the low-risk level patients,  $19.02\%$ (with $95\%$ CI: $(18.08\%, 19.96\%)$) for the medium-risk level patients and $83.89\%$ (with $95\%$ CI: $(82.93\%, 84.86\%)$) for the high-risk level patients.

\textbf{Conclusion}: Based on the census data in both communities and hospitals from Shanxi Province,   we investigate different risk factors of getting stroke and their ranking with interpretable machine learning models. The results show that Hypertension (Systolic blood pressure, Diastolic blood pressure), Physical Inactivity (Lack of sports), and Overweight (BMI) are ranked as the top three high risk factors of getting stroke in Shanxi province. The probability of getting stroke for a person can also be predicted via our machine learning model.




\section{Introduction}\label{sec:Intro}
Stroke, an acute cerebrovascular disease, is caused by brain tissue damage due to abnormal blood supply to the brain with cerebrovascular blockage. It includes hemorrhagic stroke and ischemic stroke. According to the Global Burden of Disease, Injuries and Risk Factor Study and other researches \cite{liu2007stroke, roth2018global, zhou2019mortality}, stroke is the third leading cause of death in the world and the first in China. Recent studies from National Epidemiological Survey of Stroke in China(NESS-China) \cite{wang2017prevalence} show the prevalence of stroke in China during 2012-2013:

\begin{table}[H]
\small
\centering
\scalebox{0.7}{
\begin{tabular}{cccc}
\hline
\textbf{Region} & \textbf{Prevalence (Per 100000)} & \textbf{Incidence (per 100000)} & \textbf{Mortality (per 100000)} \\ \hline
Central China   & 1549.5                         & 326.1                                & 153.7                                \\
Northeast China & 1450.3                         & 365.2                                & 158.5                                \\
South China     & 624.5                          & 154.6                                & 65                                   \\ \hline
\end{tabular}}
\caption{Prevalence of Stroke in China}
\end{table}

Investigation into risk factors of getting stroke is essentially important for the prevention of stroke. Research shows that risk factors can be divided into two categories: reversible factors and irreversible factors.

Reversible factors mainly refer to unhealthy lifestyles such as smoking, excessive alcohol consumption and physical inactivity; while irreversible factors mainly refer to chronic diseases such as hypertension, diabetes, and hyperlipidemia. A number of researches on stroke risk analysis have been done for the European and American populations \cite{vartiainen2016predicting, lumley2002stroke}. However, they can not be directly applied to the Chinese population due to racial difference. 

In China, stroke-related research is mostly carried out on risk prediction models with pathogenic factors. The most widely used one is the 10-year risk prediction model using cardiovascular and cerebrovascular diseases to give probability of stroke and coronary heart disease incidence. The CHINA-PAR project (Prediction for ASCVD Risk in China) led by Gu Dongfeng's team \cite{yang2016predicting} proposed a revised model which considered not only the 10-year risk but also a lifetime-risk assessment.  By analyzing  data on the incidence of stroke in 32 of 34 provincial regions of China, Xu et al. \cite{xu2013there}  concluded that there is  a stroke belt in north and west China. 


In recent years, some machine learning methods have been applied to the stroke prediction. In 2010, a combination of Support Vector Machine and Cox Proportional Hazard Model was proposed  by Khosla et al.  \cite{khosla2010integrated}. Benjamin \cite{letham2015interpretable} implemented an interpretable method using Decision List with Bayesian Analysis to quantify the probability of stroke. Chi-Chun Lee's team \cite{hung2017comparing, hung2019development} compared multiple methods including Deep Neural Network in stroke prediction with Electronic Health Records (EHR). In their research, they focus on the patient's 3 year stroke rate and 8 year stroke rate. However, few of these studies modeled the early screening and prevention of stroke. 

 Evaluation of the risk of getting stroke is important for prevention and treatment of stroke in China. The China National Stroke Prevention Project (CSPP)\label{project:CSPP} proposed ``8 + 2" main risk factors in identifying Chinese residents' risk level of getting stroke \cite{yu2016csdc, li2019using,chao2021stroke}:

\begin{itemize}\label{factor:main-risk-factor}
    \setlength{\itemsep}{0pt}
    \setlength{\parsep}{0pt}
    \setlength{\parskip}{0pt}
    \item[1.] Hypertension
    \item[2.] Diabetes mellitus
    \item[3.] Heart disease(includes atrial fibrillation and valvular heart disease)
    \item[4.] Hyperlipidemia
    \item[5.] Family history of stroke
    \item[6.] Overweight
    \item[7.] Smoking
    \item[8.] Physical inactivity  
    \item[\textit{a.}] The history of stroke
    \item[\textit{b.}] The history of Transient Ischemic Attack (TIA)
\end{itemize}

With the above proposed ``8+2'' main risk factors, the risk level of getting stroke can be classified into:
\begin{itemize}
  \item[1.] \textbf{High risk:} having at least three factors from factor 1 to 8; or one of \textit{a} and \textit{b};
  \item[2.] \textbf{Medium risk:} having less than three risk factors from factor 1 to 8 with at least one being factor 1, 2 or 3; 
  \item[3.] \textbf{Low risk:} having less than three risk factors from factor 4 to 8.
\end{itemize}
 
 However, the ranking of the risk factors may present differently in different provinces. Based on the census data in both communities and hospitals from Shanxi Province, in this paper, we investigates different stroke risk factors and their ranking. It shows that hypertension, physical inactivity (lack of sports), and overweight are ranked as the top three high stroke risk factors in Shanxi. The probability of getting a stroke is also estimated through our interpretable machine learning methods. The study provides theoretical support for stroke prevention and control in Shanxi Province.
\section{Materials and Methods}\label{sec:data-description}

\subsection{Dataset and Preprocessing}
 Our data is composed by two  survey datasets from 2017 to 2020:
\begin{itemize}[leftmargin=50pt]
    \setlength{\itemsep}{0pt}
    \setlength{\parsep}{0pt}
    \setlength{\parskip}{0pt}
    \item[Dataset 1:] Census in hospital: 2000 hospitalized stroke patients in 2018;
    \item[Dataset 2:] Census in community: 27583 residents during the year 2017 to 2020. This dataset is categorized and labeled with the CSPP's taxonomy: 
    \textit{Low risk (11739), Medium risk (7630) and High risk (8214).}
\end{itemize}

Each record in both datasets contains 177 features, not only providing information on the ``8+2" risk factors but also patients' other information:
\begin{table}[H]
\small
\centering
\begin{tabular}{ll}
\hline
\textbf{Feature   Name}      & \textbf{Example}                            \\ \hline
Demographic information      & Sex, Ethnicity, etc.                        \\
Lifestyle information        & Smoking, Alcohol consumption, etc.          \\
Medical measurement          & Blood pressure, fasting blood glucose, etc. \\
Surgery information          & history of surgery (PCI, CABG, CEA, CAS)    \\
Chronic diseases information & diagnosis times, what kinds of treatment    \\ \hline
\end{tabular}
\end{table}

 Data cleansing is a preparation process in data analysis by removing or correcting data that is corrupt or inaccurate.  The raw data in the above datasets needs to be cleaned due to data incompletion, inconsistence and non-structured formats which may lead to a failure of feature engineering. In this paper, missing values of a feature are filled with -1 as an abnormal class.  If there is over $60\%$ missing inside the column, we will delete it since the data from the column cannot provide much information.  Inconsistent values are found and corrected with prior medical knowledge. For instance, diastolic blood pressure should be lower than systolic blood pressure.  
 
 After the data cleansing, there are total 23289 records (low: 9718, mid: 6742, high: 5610) with 32 features in remains, shown in Table \ref{table:features}.  
\begin{table}[H]
\small
\centering
\begin{tabular}{lll}
\hline
\textbf{Class}                          & \textbf{Feature Name}    & \textbf{Data Type} \\ \hline
Lifestyle Information                   & Favor                    & Categorical        \\
Lifestyle Information                   & Alcohol                  & Categorical        \\
Lifestyle Information                   & Frequency of Vegetables  & Categorical       \\
Lifestyle Information                   & Frequency of Fruits      & Categorical       \\
Lifestyle Information                   & Meat and Vegetables      & Categorical       \\
Lifestyle Information                   & Medical Payment Method   & Categorical       \\
Demographic Information                 & Sex                      & Categorical        \\
Demographic Information                 & Age                      & Numerical          \\
Demographic Information                 & BMI                      & Numerical          \\
Demographic Information                 & Retire                   & Categorical        \\
Demographic Information                 & Height                   & Numerical          \\
Demographic Information                 & Weight                   & Numerical          \\
Demographic Information                 & Ethnicity                & Categorical        \\
Demographic Information                 & Occupation               & Categorical        \\
Demographic Information                 & Marital Status           & Categorical        \\
Demographic Information                 & Education Level          & Categorical        \\
Medical Measurement                     & TC                       & Numerical          \\
Medical Measurement                     & TG                       & Numerical          \\
Medical Measurement                     & HDL                      & Numerical          \\
Medical Measurement                     & LDL                      & Numerical          \\
Medical Measurement                     & HCY                      & Numerical          \\
Medical Measurement                     & FBG                      & Numerical          \\
Medical Measurement                     & Pulse                    & Numerical          \\
Medical Measurement                     & Systolic blood pressure  & Numerical          \\
Medical Measurement                     & Diastolic blood pressure & Numerical          \\
``8+2" Factor and Lifestyle Information  & Smoking                  & Categorical        \\
``8+2" Factor and Lifestyle Information  & Physical Inactivity      & Categorical        \\
``8+2" Factor and Medical Information    & Heart Disease            & Categorical        \\
``8+2" Factor and Medical Information    & Hypertension             & Categorical        \\
``8+2" Factor and Medical Information    & Hyperlipidemia           & Categorical        \\
``8+2" Factor and Medical Information    & History of Stroke        & Categorical        \\ 
``8+2" Factor and Medical Information    & Diabetes Melltius        & Categorical        \\
``8+2" Factor and Medical Information    & Family history of Stroke & Categorical        \\
``8+2" Factor and Medical Information    & History of Transient Ischemic Attack        & Categorical \\ \hline
\end{tabular}
\caption{Remaining features after Data Cleansing}
\label{table:features}
\end{table}

\subsection{Models}
\par\noindent\textbf{Decision-Tree} is a classic non-parametric machine learning algorithm. A tree is created through learning decision rules inferred from data features. Starting from the top root node, data are split into different internal nodes according to certain cutoff values in features, and then finally arrive the terminal leaf nodes which give the final classification result. ID3 \cite{quinlan1986induction} and CART \cite{breiman1984classification} are classic Decision-Tree algorithms which employ Information Gain and Gini Impurity from Entropy Theory \cite{pal1991entropy} as measurements in making best splitting rules.

\par\noindent\textbf{Random-Forest} is a machine learning algorithm proposed by Leo Breiman\cite{breiman2001random} in 2001. Instead of using one decision tree which is nonunique and may exhibits high variance, random forest generates a number of individual decision trees operating as a committee. Bootstrapping technique is used to train the individual decision trees in parallel on different sub datasets and features with random sampling with replacement. The final decision of classification is aggregated by voting and averaging. With the wisdom of crowds, random forest can easily overcome overfitting problem and reduce model bias caused by data imbalance, and thus shows good generalization.

\par\noindent\textbf{Logistic model} is a generalized linear model which is widely used in data mining. It assumes that the dependent variable $y$ follows a Bernoulli distribution and introduces non-linear factors through the Sigmoid function:
$$    y = \frac{1}{1 + e^{-z}},$$ where  $z = \beta_0 + \beta_1 x_1 + \dots + \beta_n x_n$  and $n$ is the number of features.

Assumes that $y$ represents a binary outcome $\{0, 1\}$, and $X$ is an array of their features and $\beta_i$ is the coefficient of feature $x_i$ \cite{peduzzi1996simulation}. The coefficient in logistic regression is called \textit{log odds} and used in logistic regression equation for the prediction of the dependent variable $y$ from the independent variable $X$, let $p = P(y=1)$:
\begin{align*}
    log\left(\frac{p}{1 - p}\right) = \beta_0 + \beta_1 x_1 + \dots + \beta_n x_n \Rightarrow p = \frac{exp(\beta_0 + \beta_1 x_1 + \dots + \beta_n x_n)}{1 + exp(\beta_0 + \beta_1 x_1 + \dots + \beta_n x_n)} \\
\end{align*}
In practice, Logistics Regression can be used in multiple aspects, for instance, advertising, disease diagnosis as it can provide the possibility of a user buying a certain product and the possibility of a certain patient suffering from a certain disease. In our case, we want to use the ``8+2" risk factors and resident's lifestyle factors as input and give out the probability of stroke incidence, which can provide a forward-looking prediction.

\subsection{Model's Interpretation}\label{Model's Interpretation}

The model's interpretability and explanations are crucial for the medical data analysis: the medical diagnosis system must be transparent, understandable, and explainable. Therefore, the doctor and the patient can know how the model makes decisions, which features are important, and how the features affect the model's result \cite{ahmad2018interpretable, molnar2020interpretable}. In this section, we mainly introduce the feature importance, permutation importance and SHAP value which can help interpret the model.
\\
\par\noindent\textbf{Feature importance}, also called as Gini importance or Mean Decrease Impurity (MDI)\cite{breiman2001random,archer2008empirical, scikit-learn}, is the average of node impurity decrease of each variable and weighted by the probability of the sample reaching to that node. For Random-Forest model, assumes that the response is $Y$ and to calculate the average variable importance of feature $X_i$ with $N$ trees:

$$Imp(X_i) = \frac{1}{N} \sum_{T=1}^{N} \sum_{j\in T: v(s_j)=X_m} p(j) \Delta i(s_j, j)$$

Where $p(j) \Delta i(s_j, j)$ is the weighted impurity decreases for feature $X_i$ in all nodes $j$, $p(j)$ is the probability of the sample reading to node ($p(j) = \frac{N_j}{N} = \frac{\text{the amount of samples reaching the node $j$}}{\text{total amount of samples}}$) and $i(s_j, j)$ is the impurity measure at node $j$ with split $s_j$. $v(s_j)$ is the variable used in the split $s_j$(split $s_j$ means the split at node $j$. Hence, $v(s_j) = X_m$ means at node $j$, the splitting identifier is variable $X_m$).

For the Decision-Tree model, it only contains one tree, that is, $N = 1$, and its feature importance can be rewritten as:

$$Imp(X_i) = \sum_{v(s_j)=X_m} p(j) \Delta i(s_j, j)$$
\\
\par\noindent\textbf{Permutation Importance} \cite{breiman2001random,altmann2010permutation, fisher2019all,scikit-learn}, is used in answering how a certain feature influence the overall prediction, as it evaluates the changes of model prediction's accuracy by permuting the feature's values. If let $s$ represent the model accuracy with the full dataset $D$, then the permutation feature importance of $i^{th}$ feature  is:

$$Per_{Imp}(i) = s - \frac{1}{K} \sum_{j=1}^K s_{i, j}, $$
where $j$ represents the $j^{th}$ repetition in $K$ times shuffling for $i^{th}$ feature  , $s_{i, j}$ as the model accuracy in modified dataset $\hat{D}_{i, j}$ with $i^{th}$ feature  shuffled. With the average changed accuracy before and after shuffling can we evaluate the importance of $i^{th}$ feature.


These two importance values can show which feature is more important, however, it is unavailable for us to know whether the feature has a positive or negative effect respect to the output.

\par\noindent\textbf{SHAP(Shapley Additive explanations)} can provide not only the importance of the features but also
show how much each feature contributes, either positively or negatively, to the target variable.
 It is a method to explain each individual prediction. This idea of SHAP value comes from the Shapley value in game theory. Shapley value tells how to fairly distribute the contributions among the features, which is the marginal contributions for each feature\cite{shap1953definition}. 

The goal of SHAP is to explain the prediction of an instance ${x}^{i}$ by computing the contribution of each feature to the prediction model. The formula of SHAP is an addictive feature attribution linear model, and it is shown below:
\begin{equation}
    {{\phi }_{i}}=\sum\limits_{S\subseteq F\backslash{\{{x}^{i}\}}}{\frac{|S|!(n-|S|-1)!}{n!}}\left[v(S\cup \{{x}^{i}\})-v(S)\right] 
\nonumber\end{equation}
 with  $ S\subseteq \{{{x}^{1}},{{x}^{2}},\cdots ,{{x}^{n}}\}\backslash {{x}^{i}},F = \{{{x}^{1}},{{x}^{2}},\cdots ,{{x}^{n}}\}.$
 
 With this method, we calculate how the feature contributes to each coalition of each decision tree model and sum them up to get the total contribution of the whole prediction model. In this equation, $F\backslash{\{{x}^{i}\}}$ represents all the possible subsets without feature ${x}^{i}$, $S$ represents the sub-feature set that did not contain the result, $v(S\bigcup \{{x}^{i}\})$ represents the model output (precision, recall or accuracy, etc.) after feature ${x}^{i}$ is added to subset $S$, $v(S)$ represents the model output using subset $S$. With the multiplication for the occurrence probability for each subsets without that feature and the output different with and without that feature, the marginal contribution of each feature ${x}^{i}$ is calculated.
 
 SHAP has three properties: local accuracy, missingness, and consistency \cite{shap2017interpretation}. Local accuracy means that when approximating the original model for a specific input $x$, local accuracy requires the explanation model to at least match the output of the model for the simplified input $x'$. Missingness means that if there is a feature missing in the sample, it does not affect the output of the model. Consistency means that when the model changes and the marginal contribution of a feature increases, the corresponding Shapley value will also increase. Therefore, it is more accurate and scientific in interpreting machine learning models due to those three characteristics.

\section{Result}\label{section:results}
\subsection{Main Risk Factors Ranking}
 Due to geographic and cultural differences, the same disease may have different manifestations in different region. We hope to find the topmost influential factors in Shanxi Province. Table \ref{table:Risk-factor-exposure} shows the proportion of each risk factor's Exposure rate and the Risk attribution (RA) based on our data:
 \footnote{
\begin{align*}
    \text{Exposure rate} &= \frac{\text{Number of patients for specific disease}}{\text{Total number in dataset}} \\ 
    \text{Risk attribution} &= \frac{\text{Incidence of a specific disease in patients has history of stroke}}{\text{Incidence of a specific disease in has no history of stroke}}
\end{align*}}
\begin{table}[H]
\small
\centering
\begin{tabular}{ccc}
\hline
\multicolumn{1}{c}{\textbf{Feature}}&\multicolumn{1}{c}{\textbf{Exposure rate}} & \multicolumn{1}{c}{\textbf{Risk attribution}}\\ \hline
Hypertension                & 0.5158  & 1.187  \\
Hyperlipidemia              & 0.3765  & 0.5846 \\
Physical Inactivity         & 0.3892  & 1.721  \\
Overweight                  & 0.3373  & 0.6844 \\
Smoking                     & 0.2058  & 1.539  \\
Family history of stroke    & 0.0999  & 1.590 \\
Diabetes Mellitus           & 0.0709  & 2.613 \\
History of stroke           & 0.0483  & NA \\
Heart Disease               & 0.0052  & 11.84 \\
History of TIA              & 0.0023  & NA  \\ \hline
\end{tabular}
\caption{Main Risk Factor Exposure rate and Risk attribution} \label{table:Risk-factor-exposure}
\end{table}

To assess the ranking of main risk factors, in the first experiment, we used the dataset 2 with the ``8+2'' factors as feature and implemented the Decision-Tree model.  Table \ref{table:DT-main-factor} shows the classification result:
\begin{table}[H]
\small
\centering
\begin{tabular}{ccccc}
\hline
\textbf{}    & \textbf{precision} & \textbf{recall} & \textbf{f1-score} & \textbf{support} \\ \hline
Low risk     & 0.9707             & 0.9950             & 0.9827            & 1999           \\
Mid risk     & 0.9711             & 0.9411          & 0.9599            & 1313           \\
High risk    & 0.9751                & 0.9650          & 0.9678            & 1345           \\
accuracy     &                    &                 & 0.9721            &                  \\
macro avg    & 0.9794             & 0.9787          & 0.9789            & 4657           \\
weighted avg & 0.9799             & 0.9796          & 0.9795            & 4657           \\ \hline
\end{tabular}
\caption{Result of Decision Tree model} \label{table:DT-main-factor}
\end{table}

\begin{figure}[H]
\centering
\begin{minipage}[t]{0.48\textwidth}
\centering
\includegraphics[width=8cm]{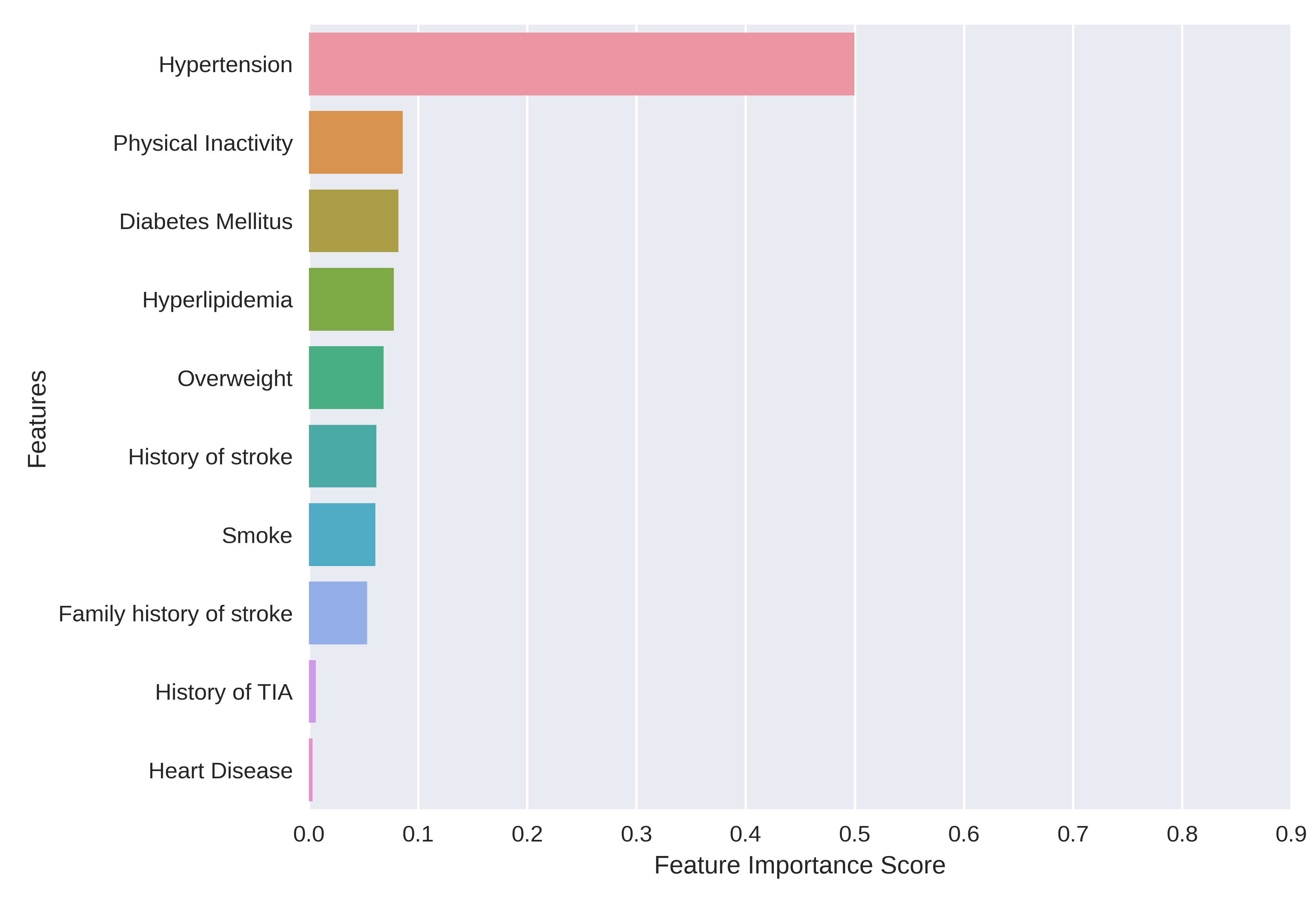}
\end{minipage}
\begin{minipage}[t]{0.48\textwidth}
\centering
\includegraphics[width=8cm]{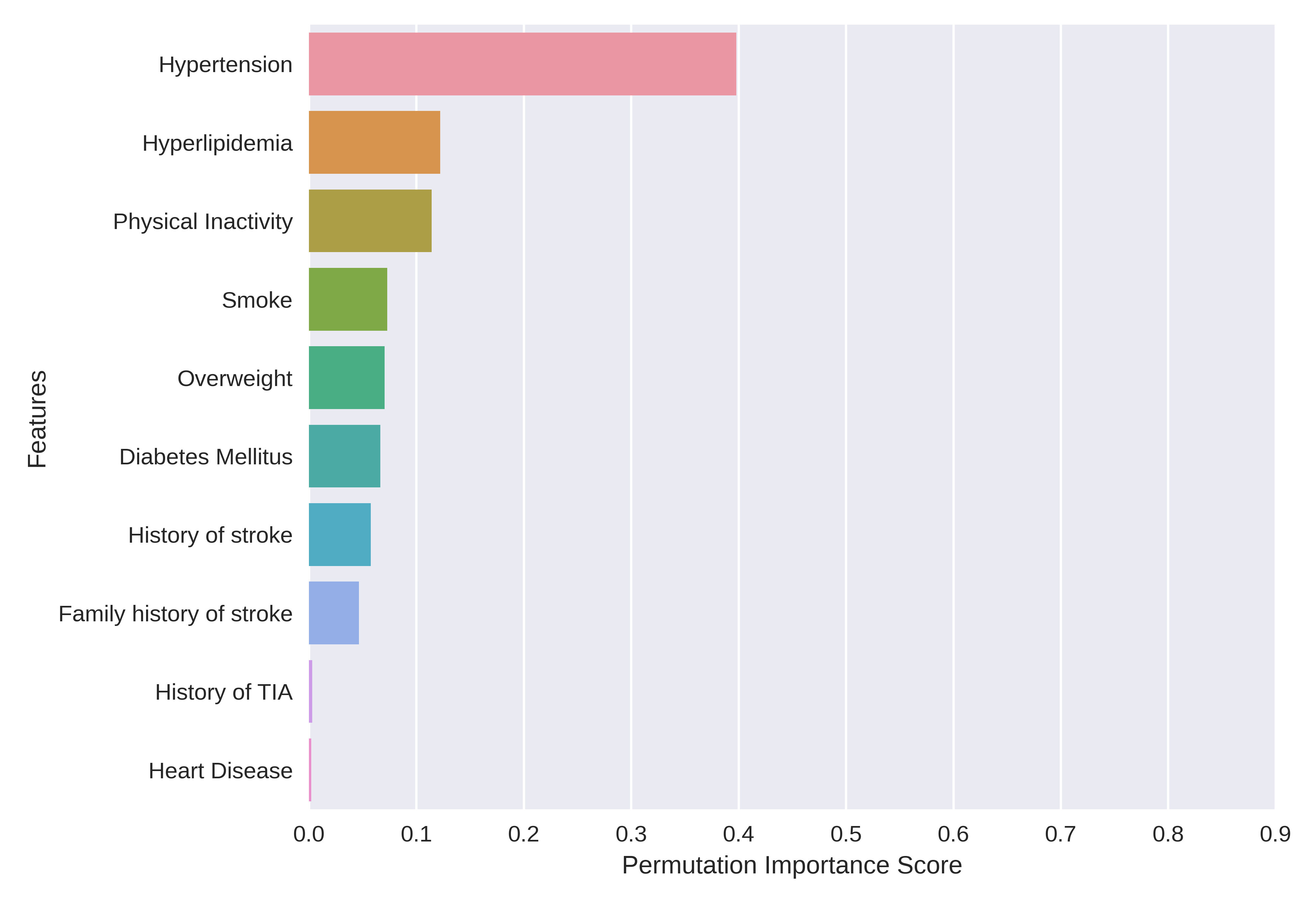}
\end{minipage}
\caption{Feature and Permutation Importance of ``8+2" Risk factor}\label{fig:feature_permutation_10_factor}
\end{figure}

Figure \ref{fig:feature_permutation_10_factor} shows the feature importance and permutation importance based on the Decision Tree model, which \textcolor{violet}{shows the ranking of these main risk factors: both evaluation methods confirm that hypertension, physical inactivity, and hyperlipidemia are estimated as the top three informative features in the Decision-Tree model.} 

\subsection{Lifestyle and Medical Measurement Ranking}
\begin{figure}[H]
\centering
\begin{minipage}[t]{0.48\textwidth}
\centering
\includegraphics[width=8cm]{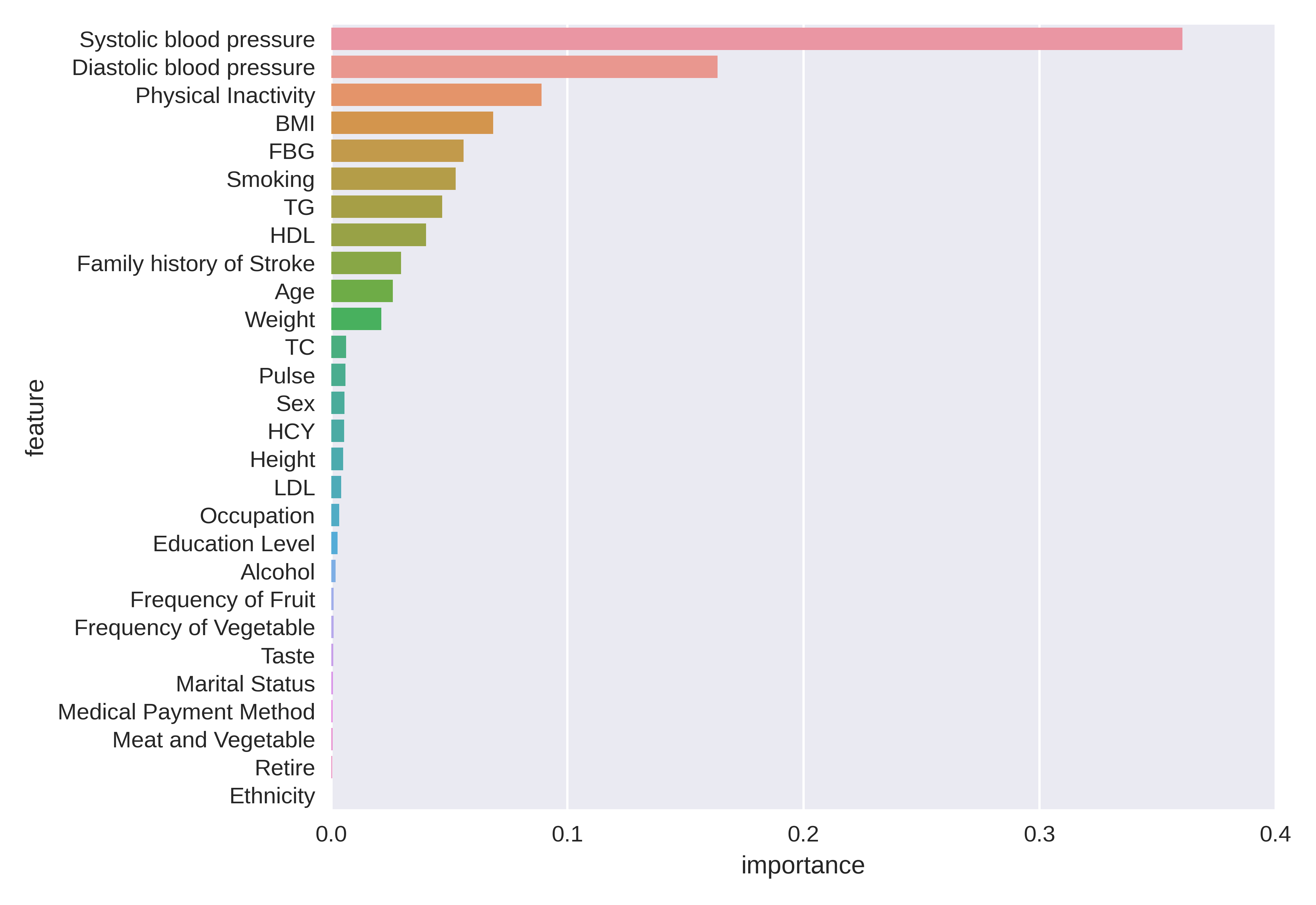}
\caption{Feature Importance}
\end{minipage}
\begin{minipage}[t]{0.48\textwidth}
\centering
\includegraphics[width=8cm]{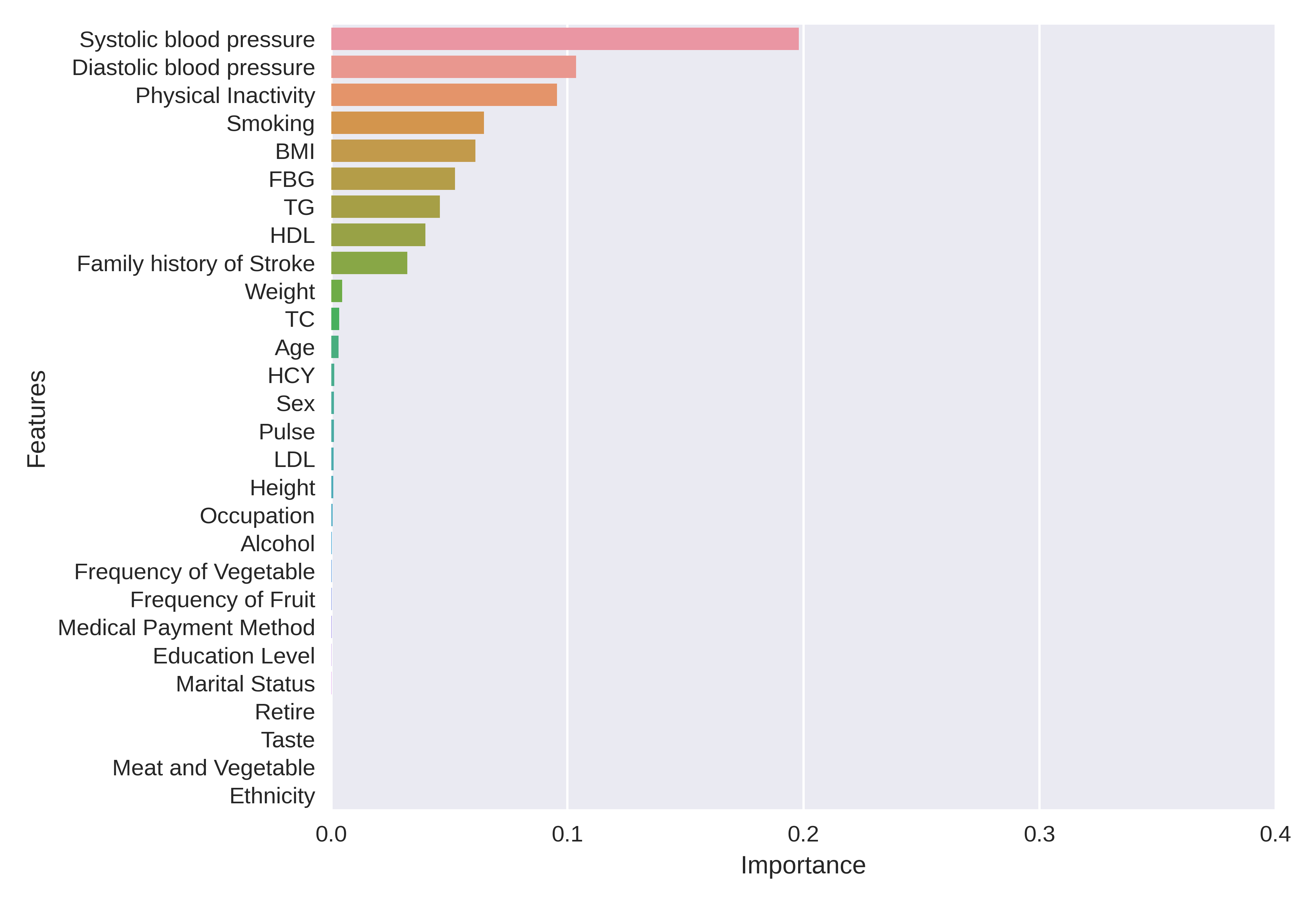}
\caption{Permutation Importance}
\end{minipage}
\caption{Feature and Permutation Importance of lifestyle and medical measurement factors factor}
\label{fig:feature_permutation_10_factor_life_med}
\end{figure}

For the second experiment, we would like to identify  
more risk factors for Shanxi Province besides the "8+2" risk factors by using the dataset 2 with features such as lifestyle habits and medical measurement.
Table \ref{table:RF-lf-mm-factors} shows the classification result and Figure \ref{fig:feature_permutation_10_factor_life_med} shows the feature and permutation importance:
\begin{table}[H]
\small
\centering
\begin{tabular}{ccccc}
\hline
\textbf{}    & \textbf{precision} & \textbf{recall} & \textbf{f1-score} & \textbf{support} \\ \hline
Low risk     & $0.8007(\pm 0.007)$             & $0.9531(\pm 0.004)$          & $0.8703(\pm 0.0041)$            & $1962$           \\
Mid risk     & $0.8213(\pm 0.008)$             & $0.7850(\pm 0.011)$          & $0.7901(\pm 0.0064)$               & $1367$           \\
High risk    & $0.9124(\pm 0.013)$             & $0.7182(\pm 0.010)$          & $0.8026(\pm 0.0076)$         & $1426$           \\
accuracy     &                    &                 & $0.84 (\pm 0.01)$            &                  \\
macro avg    & $0.8421(\pm 0.0081)$             & $0.8179(\pm 0.011)$          & $0.8271(\pm 0.0034)$            & $4755$           \\
weighted avg & $0.8311(\pm 0.0095)$             & $0.8400(\pm 0.010)$          & $0.8344(\pm 0.0062)$            & $4755$           \\ \hline
\end{tabular}
\caption{Result of Decision Tree model} \label{table:RF-lf-mm-factors}
\end{table}

 The results shown in Figure \ref{fig:feature_permutation_10_factor_life_med} confirm  that systolic blood pressure, diastolic blood pressure, physical inactivity, BMI, smoking, FBG, TG, HDL, family history of Stroke and weight are the top ten factors when we only consider the lifestyle habits, demographic information, and medical measurement. These factors are, medically, highly corresponding to the Chronic diseases \cite{levy2009genome, decode2001glucose, wu2007cut}.  

\begin{figure}[H]
    \centering
    \includegraphics[width=10cm]{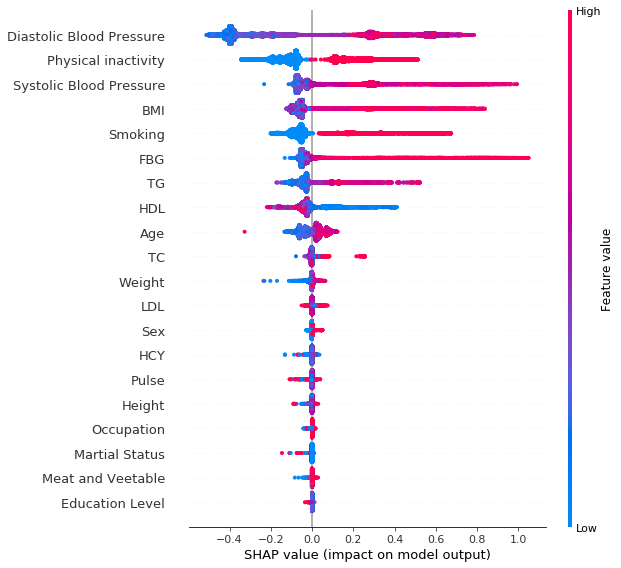}
    \caption{TreeSHAP value of top lifestyle and medical measurement factors factors}
    \label{fig:shap_value}
\end{figure}

To give out the specific details on how each feature contributes to each individual, we calculate the SHAP value in the Random Forest model and use the summary-plot to show their importance. The figure of ordered mean sample SHAP value for each feature is shown in Figure \ref{fig:shap_value}. It shows the distribution of the contributions each factor has on the cause of stroke. The color represents the feature value (red  represents  high, blue  represents  low). The more difference for the distribution between high feature value and low feature value, the better it would be in separating patients with different risk levels.  Figure \ref{fig:shap_value} shows that Diastolic Blood Pressure, Physical inactivity, Systolic Blood Pressure, BMI, Smoking, FBG, and TG are positively correlated to the risk of stroke, and HDL are negatively correlated to the risk. 

\subsection{Quantitative Prediction of Stroke's Incidence}
For the third experiment, 
a logistic model is 
 establish to  quantify the probability of stroke incidence.  To achieve this goal, we combine the dataset 1 and 2,  relabel the data: the original low-risk and medium-risk are now class 0, the high-risk and stroke are class 1. The features contains lifestyle information, demographic information, and the ``8+2" factors.

 Logistic regression is feature-sensitive.  Feature selection is done before modeling. To solve the multicollinearity problem \cite{alin2010multicollinearity},   the highly relevant features are removed first. For example, we keep BMI and remove height and weight. What's more,   Variance Threshold \cite{guyon2003introduction} is used to remove low variance features. It is a simple method of feature selection, where deletes all the features whose variance does not meet a certain threshold. For example, most of the respondent in our survey are Han Chinese, therefore, we remove ethnicity.
 
 The logistic model results are shown in Table \ref{table:logistic-factor}   including the features' coefficient, standard error, and the confidence interval. 
 According to the coefficients, it yields that History of stroke, Physical inactivity,  Hypertension, Hyperlipidemia, Smoke, Diabetes Mellitus, BMI, Family history of stroke and Heart disease are positive correlated to stroke incidence; Education level, Frequency of vegetables, and Occupation are negative correlated to stroke incidence.  
\begin{table}[H]
\small
\centering
\begin{tabular}{ccccccc}
\hline
\textbf{y=1}             & \textbf{coef} & \textbf{std err} & \textbf{z} & \textbf{$P >|z|$} & \multicolumn{2}{c}{\textbf{95\% CI {[}0.025, 0.975{]}}} \\ \hline
History of Stroke        & 2.6779  & 0.247 & 10.853  & 0.000 & 2.194  & 3.162  \\
Physical Inactivity      & 1.3948  & 0.027 & 50.815  & 0.000 & 1.341  & 1.449  \\
Hypertension             & 1.1489  & 0.027 & 41.970  & 0.000 & 1.095  & 1.203  \\
Hyperlipidemia           & 1.0875  & 0.025 & 43.210  & 0.000 & 1.038  & 1.137  \\
Smoke                    & 1.0455  & 0.031 & 33.892  & 0.000 & 0.985  & 1.106  \\
Diabetes Mellitus        & 0.8043  & 0.025 & 32.525  & 0.000 & 0.756  & 0.853  \\
BMI                      & 0.7897  & 0.026 & 30.635  & 0.000 & 0.739  & 0.840  \\
Family history of stroke & 0.7447  & 0.024 & 30.522  & 0.000 & 0.697  & 0.793  \\
Heart Disease            & 0.4257  & 0.029 & 14.630  & 0.000 & 0.369  & 0.483  \\
Frequency of Fruit       & 0.1507  & 0.026 & 5.817   & 0.000 & 0.100  & 0.202  \\
Alcohol                  & 0.1325  & 0.025 & 5.198   & 0.000 & 0.083  & 0.182  \\
Pulse                    & 0.1279  & 0.023 & 5.606   & 0.000 & 0.083  & 0.173  \\
Sex                      & 0.0863  & 0.029 & 2.945   & 0.003 & 0.029  & 0.144  \\
Retire                   & 0.0754  & 0.025 & 3.015   & 0.003 & 0.026  & 0.125  \\
Age                      & 0.0548  & 0.026 & 2.084   & 0.037 & 0.003  & 0.106  \\
Frequency of Vegetables  & -0.2262 & 0.025 & -9.118  & 0.000 & -0.275 & -0.178 \\
Occupation               & -0.2166 & 0.026 & -8.308  & 0.000 & -0.268 & -0.165 \\
Education Level          & -0.3640 & 0.027 & -13.604 & 0.000 & -0.416 & -0.312 \\
constant                 & -0.7868 & 0.068 & -11.545 & 0.000 & -0.920 & -0.653 \\ \hline
\end{tabular}
\caption{Features' Coefficient and Confidence Interval}
\label{table:logistic-factor}
\end{table}


 The trained model is implemented on the testing set  to estimate the probability of stroke incidence for each category. The results are in Table \ref{table:average_stroke_probability}:

\begin{table}[H]
\small
\centering
\begin{tabular}{cc}
\hline
\textbf{Risk   Level} & \textbf{Average Probability of Stroke} \\ \hline
Low                   & $0.0720 \ (95\% CI: (0.0665, 0.0774))$                                \\ 
Medium                & $0.1902 \ (95\% CI: (0.1808, 0.1996))$                                \\ 
High                  & $0.8389 \ (95\% CI: (0.8293, 0.8486))$                                \\ \hline
\end{tabular}
\caption{The average stroke probability for each risk level based on Logistic Model}
\label{table:average_stroke_probability}
\end{table}

Comparing with the  qualitative  ranking method, we quantify the risk factors of stroke, and convert the scoring grades into probabilities, making the prediction of stroke risk more intuitive. What's more,  our logistic model is based on the current actual circumstance to predict the incidence promptly, which is more time-sensitive.

\section{Discussion}\label{section:dicuss}
\subsection{The Risk Factors in Shanxi Province}
Based on the treeSHAP value, feature and permutation importance of lifestyle and medical measurement, we have found the most important factors of causing   stroke: 
\begin{enumerate}
    \setlength{\itemsep}{0pt}
    \setlength{\parsep}{0pt}
    \setlength{\parskip}{0pt}
    \item Hypertension(Diastolic Blood pressure and Systolic Blood Pressure)
    \item Physical Inactivity
    \item Overweight (BMI)
    \item Hyperlipidemia (mostly according to the HDL and TC)
    \item Diabetes Mellitus (according to the FBG)
\end{enumerate}

\begin{figure}[H]
\centering
\begin{minipage}[t]{0.48\textwidth}
\centering
\includegraphics[width=8cm]{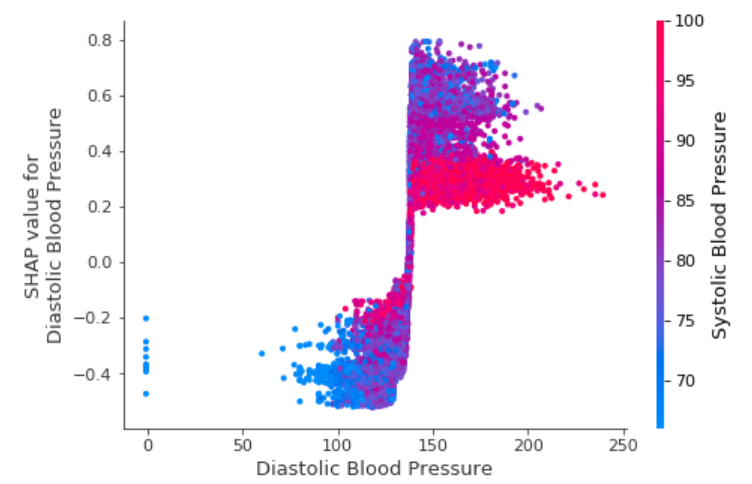}
\caption{Dependence Plot for Systolic Blood Pressure and Diastolic Blood Pressure}
\label{fig:diastolic_shap}
\end{minipage}
\begin{minipage}[t]{0.48\textwidth}
\centering
\includegraphics[width=8cm]{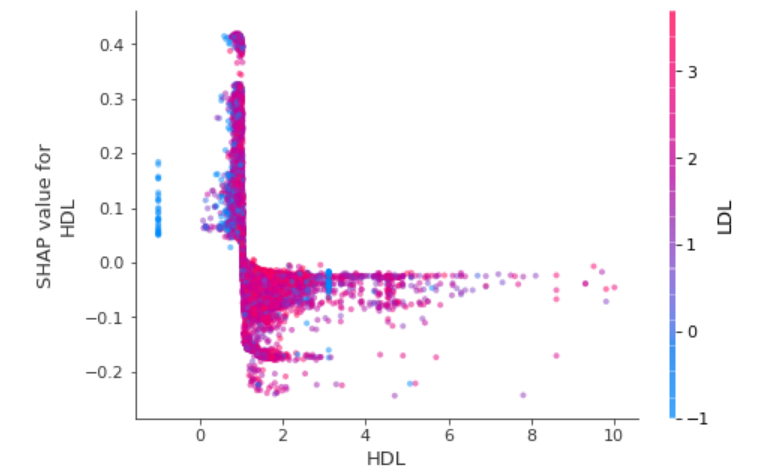}
\caption{Dependence Plot for HDL and LDL}
\label{fig:HDL_shap}
\end{minipage}
\end{figure}

 The treeSHAP dependence plot is applied to compare the contribution between two features. 
 Figure \ref{fig:diastolic_shap} shows that Diastolic Blood Pressure ($>$ 140 mmHg) is more suitable in diagnosing the risk of patients getting a stroke than the Systolic Blood Pressure ($>$ 90 mmHg). 
 Similarly, based on the comparison between HDL and LDL (see Figure \ref{fig:HDL_shap}), we can find that high-density lipoprotein are better in diagnosing those non-stroke patients in low high-density lipoprotein.

\subsection{Feature Validity}
Missing data due to the technique errors (like typos and facilities errors) is a common problem during  the census analysis. To find out how might those error data in the datasets might influence, we have conducted an experiment to find out missing data in features and its influence on the final results.  The Random Forest Classifier is adopted to predict the stroke risk  with different proportions of a single missing feature and looped for 100 times
at random locations.  What's more, to prevent the precision score didn't change due to the strong-correlation of features, some specific feature pairs are cleaned up. The result is shown in Figure \ref{fig:feature_valid} :

\begin{figure}[H]
    \centering
    \includegraphics[width=8cm]{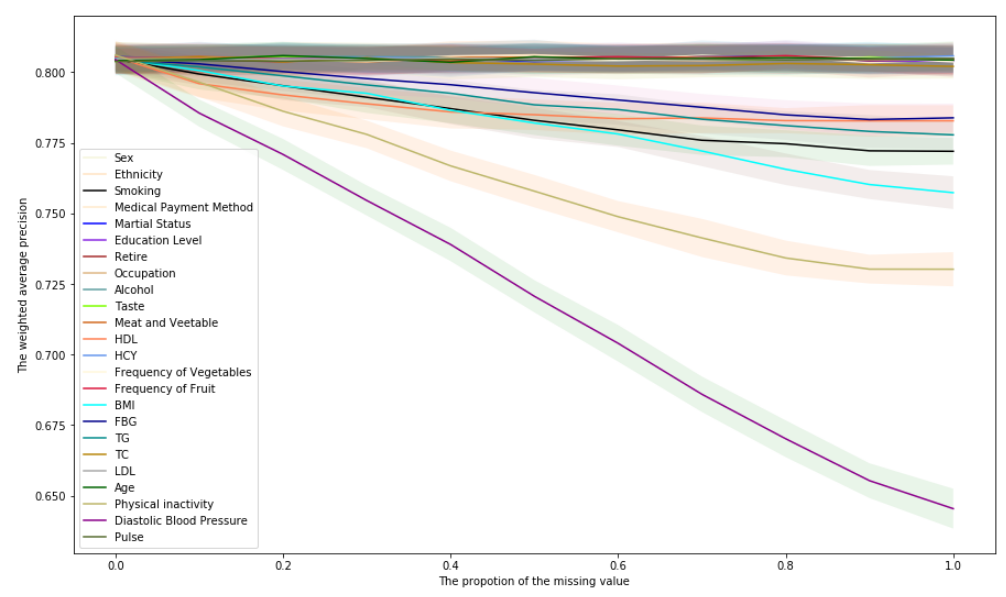}
    \caption{The relationship between the proportion of missing data in feature and the weighted precision score}
    \label{fig:feature_valid}
\end{figure}

In Figure \ref{fig:feature_valid}, the curve for each feature represents how the average weight precision score of the feature has changed with the increasing proportion of the feature missing and the shadows are the 95\% confidence area 100 times for each feature. Based on the result, we can see that diastolic blood pressure, physical inactivity, BMI, smoking, alcohol, HDL, and FBG are in order important factors when identifying the cause of stroke, while the other factors are not influencing the whole models. An interesting fact we have seen is that HDL seems to be a great influential factor when the proportion of HDL is small compared to most of the influenced factors.

\begin{figure}[H]
    \centering
    \includegraphics[width=10cm]{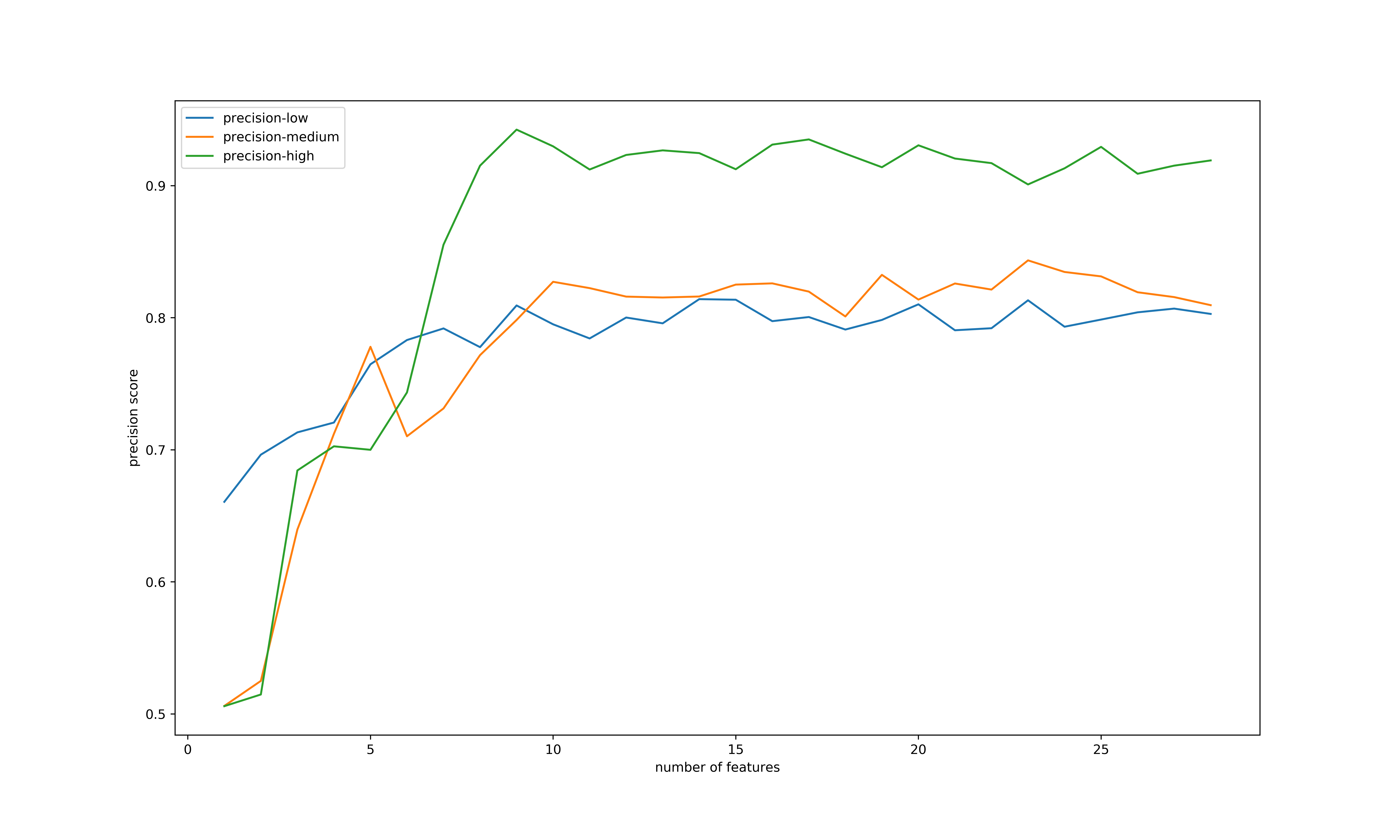}
    \caption{The relationship between the amount of features and the precision changed}
    \label{fig:feature_valid_precision}
\end{figure}

What's more, a Recursive Feature Elimination (RFE) process is also done to evaluate the specific amount of factors for analyzing the risk level of patients getting a stroke. The procedure of the RFE is as follows. First, the estimator is trained on the initial feature set and the importance of each feature is obtained from any specific or callable attribute. Then, the least important features are removed from the current feature set. This process is repeated recursively over the pruning set until the number of features to be selected is finally reached. Based on the Figure \ref{fig:feature_valid_precision}, we have found that approximate 7 features can help the Random Forest model to get a stable precision for different levels of risk. Therefore, the validity of those features is proved.

\bibliographystyle{unsrt}
\bibliography{reference}
\end{document}